\documentclass[%
 reprint,
 amsmath,amssymb,
 aps, 
 pre,
 showkeys
]{revtex4-1}

\usepackage{graphicx}
\usepackage{dcolumn}
\usepackage{bm}

\usepackage{algpseudocode, algorithm}

\usepackage{my-style}

\allowdisplaybreaks[4]

\begin{document}

\preprint{APS/123-QED}

\title{Linear-Time Algorithm in Bayesian Image Denoising based on Gaussian Markov Random Field}

\author{Muneki Yasuda}  
\affiliation{%
 Graduate School of Science and Engineering, Yamagata University.
}%
\author{Junpei Watanabe}  
\affiliation{%
 Graduate School of Science and Engineering, Yamagata University.
}%
\author{Shun Kataoka}  
\affiliation{%
 Graduate School of Information Sciences, Tohoku University.
}%
\author{kazuyuki Tanaka}  
\affiliation{%
 Graduate School of Information Sciences, Tohoku University.
}%


\begin{abstract}
In this paper, we consider Bayesian image denoising based on a Gaussian Markov random field (GMRF) model, 
for which we propose an new algorithm. 
Our method can solve Bayesian image denoising problems, including hyperparameter estimation, in $O(n)$-time, where $n$ is the number of pixels in a given image. 
From the perspective of the order of the computational time, this is a state-of-the-art algorithm for the present problem setting. 
Moreover, the results of our numerical experiments we show our method is in fact effective in practice.
\end{abstract}

\pacs{Valid PACS appear here}
\keywords{Bayesian image denoising, Gaussian Markov random field, EM algorithm, mean-field method, linear-time algorithm}
\maketitle


\section{Introduction} 
\label{sec:Introduction}

Bayesian image processing~\cite{Geman&Geman1984} based on a probabilistic graphical model has a long and rich history~\cite{MRF&CV2011}. 
In Bayesian image processing, one constructs a posterior distribution and then infers restored images based on the posterior distribution. 
The posterior distribution is derived from a prior distribution that captures the statistical properties of the images. 
One of the major challenges of Bayesian image processing is the construction of an effective prior for the images. 
For this purpose, a Gaussian Markov random field (GMRF) model (or Gaussian graphical model) is a possible choice. 
Because a GMRF is a multi-dimensional Gaussian distribution, its mathematical treatment is tractable. 
GMRFs have also been applied in various research fields other than Bayesian image processing, e.g., traffic reconstruction~\cite{Kataoka2014}, sparse modeling~\cite{GLasso2007}, and earth science~\cite{Kuwatani2014a,Kuwatani2014b}.

In this paper, we focus on Bayesian image denoising based on GMRFs~\cite{Nishimori2000,TanakaReview2002}. 
The procedure of Bayesian image denoising is divided into two stages: a hyperparameter-estimation stage and a restoring stage. 
In the hyperparameter-estimation stage, one estimates the optimal values of hyperparameters in posterior distribution for a given degraded image 
by using the expectation-maximization (EM) algorithm (or maximum marginal-likelihood estimation),  
while, in the restoring stage, one infers the restored image based on the posterior distribution with the optimal hyperparameters by using maximum a posteriori (MAP) estimation. 
However, the processes in the two stages have a computational problem. 
They require the an inverse operation of $n \times n$ covariance matrix, the computational time of which is $O(n^3)$, 
where $n$ is the number of pixels in the degraded image. 
Hence, in order to reduce the computational time, some approximations are employed. 
In the usual setting in Bayesian image denoising, the graph structure of the GMRF model is a square grid graph according to the configuration of the pixels.
By adding the periodic boundary condition (PBC) to the square grid graph, the authors of \cite{Nishimori2000,TanakaReview2002} constructed $O(n^2)$-time algorithms.
We refer to this approximation as \textit{torus approximation}, because a square grid graph with the PBC is a torus graph.  
This approximation allows the graph Laplacian of the GMRF model to be diagonalized by using discrete Fourier transformation (DFT), 
and the diagonalization can reduce the computational time to $O(n^2)$. 
More recently, it was further reduced to $O(n \ln n)$ by using fast Fourier transformation (FFT)~\cite{Nicolas2010}. 

Recently, it was shown that the graph Laplacian of a square-grid GMRF model without the PBC can be diagonalized by using discrete cosine transformation (DCT)~\cite{SDCT2015,SDCT2017,SDCT2017-Okada}. 
By using DCT, one can diagonalize the graph Laplacian without the torus approximation.
Therefore, by combining the diagonalization based on DCT with the FFT method proposed in Ref~\cite{Nicolas2010}, 
one can construct an $O(n \ln n)$-time algorithm without the torus approximation. 
The details of this algorithm are presented in \ref{app:Nicola'sMethod}. 

In this paper, we define a model for Bayesian image denoising based on GMRFs. 
The defined GMRF model has a more general form than the GMRF models presented in many previous studies, 
and it includes them as special cases. 
In our GMRF model, we propose an $O(n)$-time algorithm for practical Bayesian image denoising that uses DCT and the mean-field approximation. 
Since the mean-field approximation is equivalent to the exact computation in the present setting (see \ref{appendix:MFE}), 
the results obtained by our method are equivalent to those obtained by exact computation. 
Our algorithm is state-of-the-art algorithm in the perspective of the order of the computational time. 
It is noteworthy that, in the present problem setting, one cannot create a sublinear-time algorithm, i.e., an algorithm having a computational time is less than $O(n)$, 
because at least $O(n)$-time computation is needed to create the restored image. 
In Tab. \ref{tab:TimeComparison}, the computational times of our method and the previous methods are shown.
\begin{table}[tb]
\caption{Order of computational time. The DCT--FFT method is presented in \ref{app:Nicola'sMethod}.}
\begin{tabular}{|c|c|c|} \hline
   & Time& Approximation \\ \hline \hline
Naive method & $O(n^3)$ & exact \\ \hline
DFT method~\cite{Nishimori2000,TanakaReview2002}     & $O(n^2)$ & torus approximation  \\ \hline
DFT--FFT method~\cite{Nicolas2010}     & $O(n \ln n)$ & torus approximation  \\ \hline
DCT--FFT method     & $O(n \ln n)$ & exact  \\ \hline
Proposed method     & $O(n)$ & exact \\ \hline
\end{tabular}
\label{tab:TimeComparison}
\end{table}

The remainder of this paper is organized as follows. 
The model definition and the frameworks of Bayesian image denoising and of the EM algorithm for hyperparameters estimation are presented in Sect. \ref{sec:FrameworkBayesianDenoising}. 
The proposed method is shown in Sect. \ref{sec:LinearEM-Algorithm}. 
In Sect. \ref{sec:NumericalExperiment}, we show the validity of our method by the results of numerical experiments. 
Finally, conclusions are given in Sect. \ref{sec:conclusion}.

\section{Framework of Bayesian Image Denoising}
\label{sec:FrameworkBayesianDenoising}

\begin{figure}[tb]
\centering
\includegraphics[height=4.5cm]{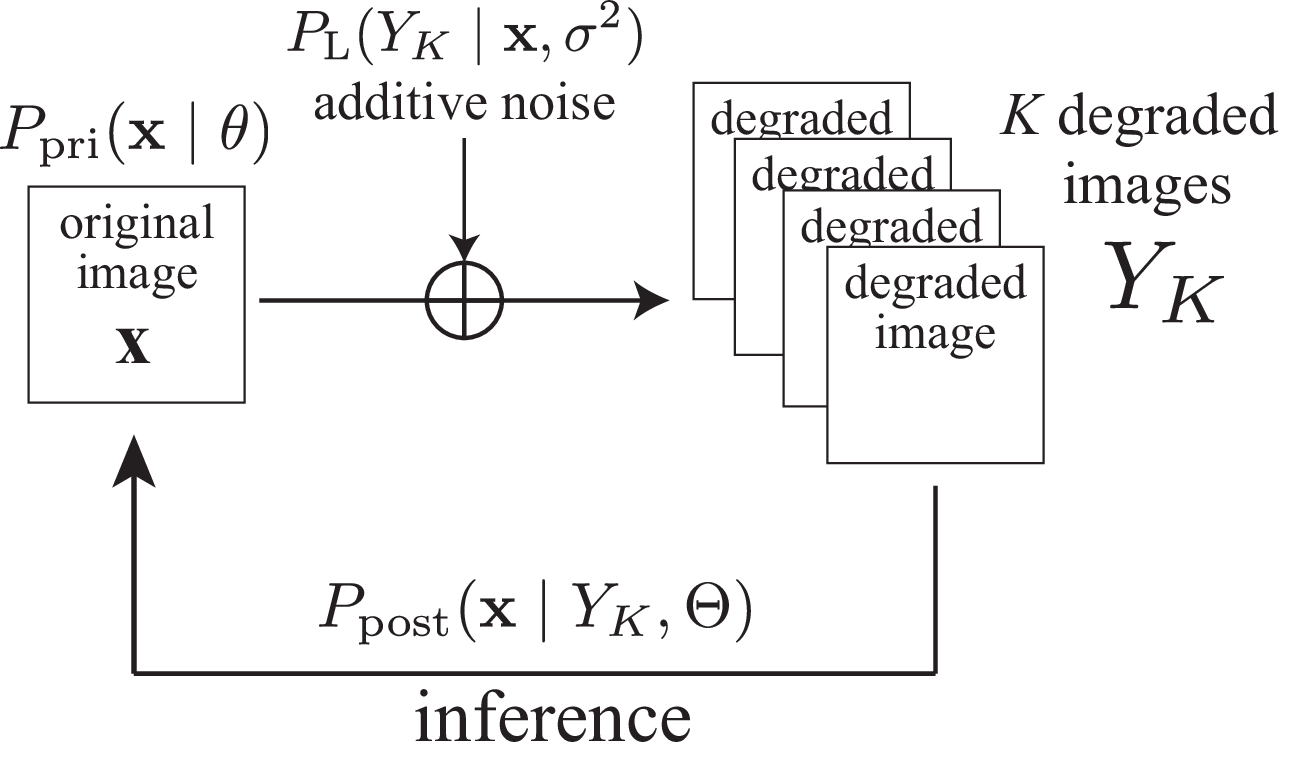}
\caption{Framework of Bayesian image denoising considered in this paper.
$P_{\mrm{pri}}(\ve{x} \mid \theta)$, $P_{\mrm{L}}(Y_K \mid \ve{x}, \sigma^2)$, and $P_{\mrm{post}}(\ve{x} \mid Y_K, \Theta)$ are the prior distribution (Eq. (\ref{eqn:prior_dist})), the likelihood (Eq. (\ref{eqn:likelihood_dist})), and the posterior distribution (Eq. (\ref{eqn:posterior_dist})), respectively. $Y_K$ is the set of $K$ degraded images.}
\label{fig:BayesFramework}
\end{figure}

Consider the problem of digital image denoising described below. 
Given an original image composed of $v \times v$ pixels, a degraded image of the same size of the original one is generated by adding additive white Gaussian noise (AWGN) to the original one. 
Suppose that we independently obtain $K$ degraded images generated from the same stochastic process, where $K\geq 1$.
The goal of Bayesian image denoising is to infer the original image from the given data, i.e., the $K$ degraded images.
We denote the original image by the vector $\ve{x} = (x_1, x_2,\ldots, x_n)^{\mrm{t}}$ ($x_i$ expresses the intensity of the $i$th pixel), 
where the image is vectorized by raster scanning (or row-major scanning) on the image, and $n = v^2$ is the number of pixels in the image;  
we denote the $k$th degraded image by the vector $\ve{y}^{(k)} =  (y_1^{(k)}, y_2^{(k)},\ldots, y_n^{(k)})^{\mrm{t}}$.
Note that the degraded images are also vectorized by the same procedure, i.e., raster scanning, as the original one. 
The framework of the presented Bayesian image denoising is illustrated in Fig. \ref{fig:BayesFramework}.

We define the prior distribution of the original image by using a GMRF model, 
defined on a $v \times v$ square grid graph $G(V, E)$ with a free boundary condition, 
and express it as
\begin{align}
P_{\mrm{pri}}(\ve{x} \mid \theta):=\frac{1}{Z_{\mrm{pri}}(\theta)} \exp\big( -E_{\mrm{pri}}(\ve{x} ; \theta)\big)
\label{eqn:prior_dist}
\end{align}
with the energy function
\begin{align}
E_{\mrm{pri}}(\ve{x} ; \theta) &:= - b \sum_{i \in V}x_i + \frac{\lambda}{2} \sum_{i \in V}x_i^2 \nn
\aleq
+\frac{\alpha}{2} \sum_{\{i,j\} \in E} (x_i -x_j)^2, 
\label{eqn:energy_prior}
\end{align}
where $V = \{1,2,\ldots, n\}$ and $E = \{\{i,j\} \}$ are the sets of vertices and undirected edges in $G(V, E)$, respectively, 
and $\theta =\{ b, \lambda, \alpha\}$ is the set of the hyperparameters in the prior distribution: 
$b$ controls the brightness of the image, $\lambda$ controls the variance of the intensities of pixels, and $\alpha$ controls the smoothness of the image. 
$Z_{\mrm{pri}}(\theta)$ is the partition function defined by
\begin{align}
Z_{\mrm{pri}}(\theta):=\int_{-\infty}^{\infty} \exp\big( -E_{\mrm{pri}}(\ve{x} ; \theta)\big) d\ve{x}.
\label{eqn:partition-function_prior}
\end{align} 
The energy function of the prior distribution in Eq. (\ref{eqn:energy_prior}) can be rewritten as 
\begin{align}
E_{\mrm{pri}}(\ve{x} ; \theta) = - \ve{b}^{\mrm{t}} \ve{x} + \frac{1}{2}  \ve{x}^{\mrm{t}} \ve{S}_{\mrm{pri}} \ve{x},
\label{eqn:energy_prior_matrix}
\end{align}
where $\ve{b} \in \mathbb{R}^{n}$ is the $n$-dimensional vector, the elements of which are all $b$, $\ve{b} = (b, b, \ldots, b)^{\mrm{t}}$, and
\begin{align}
\ve{S}_{\mrm{pri}} := \lambda \ve{I}_n + \alpha \ve{\Lambda}
\label{eqn:precision-Mat_prior}
\end{align} 
is the precision matrix (the inverse of the covariance matrix) of the prior distribution, 
where $\ve{I}_n \in \mathbb{R}^{n \times n}$ is the $n$-dimensional identity matrix and $\ve{\Lambda} \in \mathbb{R}^{n \times n}$ is the graph Laplacian of $G(V, E)$: 
\begin{align}
\Lambda_{i,j} =
\begin{cases}
|\partial(i)| & i = j\\
-1 & \{i,j\} \in E \\
0 & \mrm{others}
\end{cases}
,
\label{eqn:GraphLaplacian}
\end{align} 
where $\partial(i) \subset V$ is the set of vertices connected to vertex $i$. 
By using Eq. (\ref{eqn:energy_prior_matrix}) and the Gaussian integral, Eq. (\ref{eqn:partition-function_prior}) yields
\begin{align}
Z_{\mrm{pri}}(\theta)=\exp\Big( \frac{1}{2}  \ve{b}^{\mrm{t}} \ve{S}_{\mrm{pri}}^{-1} \ve{b}\Big)\sqrt{ (2\pi)^n \det \ve{S}_{\mrm{pri}}^{-1}}.
\label{eqn:partition-function_Mat_prior}
\end{align}
It is noteworthy that the hyperparameter $\lambda$ is important for the mathematical treatment for the prior distribution, 
because, when $\lambda = 0$, the precision matrix $\ve{S}_{\mrm{pri}}$ is not positive definite 
and therefore Eq. (\ref{eqn:prior_dist}) is not a well-defined probabilistic distribution. 

Since we assume the AWGN, the likelihood, i.e., the generating process of the degraded images, for the $K$ degraded images $Y_K = \{ \ve{y}^{(k)} \mid k = 1,2,\ldots, K\}$ is defined by
\begin{align}
&P_{\mrm{L}}(Y_K \mid \ve{x}, \sigma^2)\nn
&:= \prod_{k = 1}^K \prod_{i \in V} \frac{1}{\sqrt{2\pi \sigma^2}} \exp\Big(-\frac{(x_i - y_i^{(k)})^2}{2 \sigma^2}\Big),
\label{eqn:likelihood_dist}
\end{align}
where $\sigma^2 > 0$ is the variance of the AWGN.

By using Eqs. (\ref{eqn:prior_dist}) and (\ref{eqn:likelihood_dist}) and Bayesian theorem 
\begin{align*}
P_{\mrm{post}}(\ve{x} \mid Y_K, \Theta) \propto P_{\mrm{L}}(Y_K \mid \ve{x}, \sigma^2)P_{\mrm{pri}}(\ve{x} \mid \theta),
\end{align*}
where $\Theta = \theta \cup\{ \sigma^2\}=\{ \sigma^2, b, \lambda, \alpha\}$, 
the posterior distribution is expressed as
\begin{align}
&P_{\mrm{post}}(\ve{x} \mid Y_K, \Theta)\nn
&:=\frac{1}{Z_{\mrm{post}}(\Theta)}\exp\Big( -\frac{1}{2\sigma^2}\sum_{k =1}^K || \ve{x}- \ve{y}^{(k)} ||^2\nn
\aleq
 -E_{\mrm{pri}}(\ve{x} ; b, \lambda, \alpha)\Big),
\label{eqn:posterior_dist}
\end{align}
where
\begin{align}
Z_{\mrm{post}}(\Theta)
&:=\int_{-\infty}^{\infty} \exp\Big( -\frac{1}{2\sigma^2}\sum_{k =1}^K || \ve{x}- \ve{y}^{(k)} ||^2 \nn
\aleq 
-E_{\mrm{pri}}(\ve{x} ; b, \lambda, \alpha)\Big) d\ve{x}.
\label{eqn:partition-function_post}
\end{align}
Here, $|| \cdots ||$ is the Euclidean norm of the assigned vector.
Similarly to the prior distribution, the posterior distribution and its partition function in Eqs. (\ref{eqn:posterior_dist}) and (\ref{eqn:partition-function_post}) can be expressed as
\begin{align}
&P_{\mrm{post}}(\ve{x} \mid Y_K, \Theta)\nn
&=\frac{1}{Z_{\mrm{post}}(\Theta)}\exp\Big(-\frac{1}{2\sigma^2}\sum_{k = 1}^K ||\ve{y}^{(k)}||^2
+ \ve{c}^{\mrm{t}} \ve{x} \nn
\aleq
- \frac{1}{2}  \ve{x}^{\mrm{t}} \ve{S}_{\mrm{post}} \ve{x}\Big)
\label{eqn:posterior_dist_Mat}
\end{align}
and
\begin{align}
Z_{\mrm{post}}(\Theta) &= \exp\Big(-\frac{1}{2\sigma^2}\sum_{k = 1}^K ||\ve{y}^{(k)}||^2 + \frac{1}{2}  \ve{c}^{\mrm{t}} \ve{S}_{\mrm{post}}^{-1} \ve{c}\Big)\nn
\aleq
\times
\sqrt{ (2\pi)^n \det \ve{S}_{\mrm{post}}^{-1}},
\label{eqn:partition-function_Mat_post}
\end{align}
respectively, 
where 
\begin{align}
\ve{c} := \ve{b} + \frac{K}{\sigma^2}\hat{\ve{y}},
\end{align}
where
\begin{align}
\hat{\ve{y}} := \frac{1}{K}\sum_{k = 1}^K \ve{y}^{(k)}
\label{eqn:average_y_k}
\end{align}
is the average image of $K$ degraded images, and
\begin{align}
\ve{S}_{\mrm{post}}:= \Big(\lambda + \frac{K}{\sigma^2}\Big) \ve{I}_n + \alpha \ve{\Lambda}
\label{eqn:precision-Mat_post}
\end{align}
is the precision matrix of the posterior distribution.

In previous work~\cite{Nishimori2000,TanakaReview2002,Nicolas2010}, $b$ and $\lambda$ were treated as fixed constants, 
whereas, in this study, they were treated as controllable parameters, which are optimized through the EM algorithm as explained later. 
Moreover, the previous work considered only the case where the number of degraded images is just one, i.e., the case of $K = 1$, 
while the present formulation allows the treatment of multiple degraded images. 
For these reasons, our model is a generalization model that includes the models presented in the previous studies~\cite{Nishimori2000,TanakaReview2002,Nicolas2010}.

\subsection{Maximum a Posteriori Estimation}
\label{sec:MAP-estimation}

In Bayesian image denoising, we have to infer the original image by using only the $K$ degraded images $Y_K$. 
In MAP estimation, we regard that $\ve{m}(Y_K, \Theta) = (m_1(Y_K, \Theta), m_2(Y_K, \Theta),\ldots, m_n(Y_K, \Theta))^{\mrm{t}}$,  
which maximize $P_{\mrm{post}}(\ve{x} \mid Y_K, \Theta)$, 
\begin{align}
\ve{m}(Y_K, \Theta) := \argmax_{\ve{x}} P_{\mrm{post}}(\ve{x} \mid Y_K, \Theta),
\label{eqn:MAP_estimation}
\end{align}
is the most probable as the estimation of the original. 
Since the posterior distribution in Eq. (\ref{eqn:posterior_dist}) is multivariate Gaussian distribution, 
the MAP estimate $\ve{m}(Y_K, \Theta)$ coincides with the mean vector of the posterior distribution: 
\begin{align*}
\ve{m}(Y_K, \Theta) = \int_{-\infty}^{\infty} \ve{x} P_{\mrm{post}}(\ve{x} \mid Y_K, \Theta) d\ve{x}.
\end{align*}
Naively, the order of the computational time of computing the mean vector of the posterior distribution in Eq. (\ref{eqn:MAP_estimation}) is $O(n^3)$, 
because it includes the inverting operation for the precision matrix: $\ve{m}(Y_K, \Theta) = \ve{S}_{\mrm{post}}^{-1} \ve{c}$.

We can obtain linear equations among $\ve{m}(Y_K, \Theta)$ by using mean-field approximation~\cite{GraphicalModel2008} or loopy belief propagation~\cite{GaBP2001}. 
Both methods lead to the same expression: 
\begin{align}
m_i(Y_K, \Theta) &= \frac{1}{\lambda + \alpha |\partial(i)| + K / \sigma^2}
\Big( b + \frac{K}{\sigma^2}\hat{y}_i \nn
\aleq
+ \alpha \sum_{j \in \partial(i)} m_j(Y_K, \Theta)\Big).
\label{eqn:GS-method_post}
\end{align}
Eq. (\ref{eqn:GS-method_post}) can be solved by using a successive iteration method. 
It is known that, in a GMRF model, the mean vector evaluated by the mean-field approximation (or the loopy belief propagation) is equivalent to the exact one~\cite{GaBP2001,GraphicalModel2008,GaBP2008};  
see \ref{appendix:MFE}.
Therefore, the solution to Eq. (\ref{eqn:GS-method_post}) is equivalent to the exact mean vector of the posterior distribution in Eq. (\ref{eqn:MAP_estimation}). 
The order of the computational time of solving Eq. (\ref{eqn:GS-method_post}) is $O(n)$, 
because the second sum in the right hand side of Eq. (\ref{eqn:GS-method_post}) includes at most four terms.

From the above, we see that, for a given $Y_K$ and a fixed $\Theta$, Bayesian image denoising can be performed in $O(n)$-time.
Obviously, however, the quality of the result should depend on $\Theta$. 
The stage of optimization of $\Theta$ still remains. 
The EM algorithm, described in the following section, is one of the most popular methods for the optimization of $\Theta$.

\subsection{Expectation-Maximization Algorithm}
\label{sec:EM-Algorithm}

In the EM algorithm, for a fixed $\Theta_{\mrm{old}} = \theta_{\mrm{old}} \cup\{ \sigma_{\mrm{old}}^2\}= \{ \sigma_{\mrm{old}}^2, b_{\mrm{old}}, \lambda_{\mrm{old}}, \alpha_{\mrm{old}}\}$, we have to maximize the $Q$-function, 
\begin{align}
&Q(\Theta ; \Theta_{\mrm{old}})\nn
&:= \int_{-\infty}^{\infty} P_{\mrm{post}}(\ve{x} \mid Y_K, \Theta_{\mrm{old}}) \ln P_{\mrm{joint}}(\ve{x} ,Y_K \mid \Theta) d\ve{x},
\label{eqn:Q-function}
\end{align}
with respect to $\Theta$, where $P_{\mrm{joint}}(\ve{x}, Y_K \mid \Theta) := P_{\mrm{L}}(Y_K \mid \ve{x}, \sigma^2)P_{\mrm{pri}}(\ve{x} \mid \theta)$ is the joint distribution over $\ve{x}$ and $Y_K$.
The gradients of the $Q$-function for the parameters in $\Theta$ are
\begin{align}
\nabla_b Q(\Theta ; \Theta_{\mrm{old}}) &= \sum_{i \in V} \mathbb{E}_{\mrm{post}} [ x_i \mid \Theta_{\mrm{old}}]\nn
\aleq - \sum_{i \in V} \mathbb{E}_{\mrm{pri}}[x_i \mid \theta],
\label{eqn:grad-b}\\
\nabla_{\lambda} Q(\Theta ; \Theta_{\mrm{old}}) &= -\frac{1}{2} \mathbb{E}_{\mrm{post}} \big[ || \ve{x} ||^2 \mid \Theta_{\mrm{old}} \big]\nn
\aleq
+ \frac{1}{2} \mathbb{E}_{\mrm{pri}} \big[|| \ve{x} ||^2 \mid \theta \big], 
\label{eqn:grad-lambda}\\
\nabla_{\sigma^2} Q(\Theta ; \Theta_{\mrm{old}})& = \frac{1}{2 \sigma^4} \sum_{k = 1}^K \mathbb{E}_{\mrm{post}} \big[|| \ve{x} - \ve{y}^{(k)} ||^2 \mid \Theta_{\mrm{old}} \big] \nn
\aleq
- \frac{n K}{2\sigma^2},
\label{eqn:grad-sigma}
\end{align}
and
\begin{align}
\nabla_{\alpha}Q(\Theta ; \Theta_{\mrm{old}}) &= -\frac{1}{2}\sum_{\{i,j\} \in E} \mathbb{E}_{\mrm{post}}[(x_i- x_j)^2 \mid \Theta_{\mrm{old}}]\nn
\aleq
+ \frac{1}{2}\sum_{\{i,j\} \in E} \mathbb{E}_{\mrm{pri}}[(x_i- x_j)^2 \mid \theta],
\label{eqn:grad-alpha}
\end{align}
where $\nabla_{\gamma} Q(\Theta ; \Theta_{\mrm{old}}) := \partial Q(\Theta ; \Theta_{\mrm{old}}) / \partial \gamma$, 
and $\mathbb{E}_{\mrm{pri}}[f(\ve{x}) \mid \theta]$ and $\mathbb{E}_{\mrm{post}}[f(\ve{x}) \mid \Theta]$ are the expectation values of $f(\ve{x})$ with respect to 
the prior distribution $P_{\mrm{pri}}(\ve{x} \mid \theta)$ and the posterior distribution $P_{\mrm{post}}(\ve{x} \mid Y_K, \Theta)$, respectively. 
Using a gradient ascent method with the gradients in Eqs. (\ref{eqn:grad-b})--(\ref{eqn:grad-alpha}), we maximize the $Q$-function in Eq. (\ref{eqn:Q-function}). 
Naively, the computation of these gradients requires $O(n^3)$-time, which is rather expensive, 
because they include the inverting operation for the precision matrices, $\ve{S}_{\mrm{pri}}$ and $\ve{S}_{\mrm{post}}$. 

The method presented in the next section allows the EM algorithm to be executed in a linear time. 
Therefore, given $K$ degraded images, we can perform Bayesian image denoising, including the optimization of $\Theta$ by the EM algorithm, in $O(n)$-time.

\section{Expectation-Maximization Algorithm in Linear Time}
\label{sec:LinearEM-Algorithm}

We now consider the free energies of the prior and the posterior distributions,
\begin{align}
F_{\mrm{pri}}(\theta)&:= -\ln Z_{\mrm{pri}}(\theta),
\label{eqn:def_free-energy_prior}\\
F_{\mrm{post}}(\Theta) &:= -\ln Z_{\mrm{post}}(\Theta).
\label{eqn:def_free-energy_post}
\end{align}
By using these free energies, the gradients in Eqs. (\ref{eqn:grad-b})--(\ref{eqn:grad-alpha}) are rewritten as
\begin{align}
\nabla_b Q(\Theta ; \Theta_{\mrm{old}}) &= \frac{\partial F_{\mrm{pri}}(\theta)}{\partial b} -\frac{\partial F_{\mrm{post}}(\Theta_{\mrm{post}})}{\partial b}, 
\label{eqn:grad-b_freeEnergyExpression}\\
\nabla_{\lambda} Q(\Theta ; \Theta_{\mrm{old}}) &= \frac{\partial F_{\mrm{pri}}(\theta)}{\partial \lambda} - \frac{\partial  F_{\mrm{post}}(\Theta_{\mrm{old}})}{\partial \lambda_{\mrm{old}}}, 
\label{eqn:grad-lambda_freeEnergyExpression}\\
\nabla_{\sigma^2} Q(\Theta ; \Theta_{\mrm{old}})& = -\frac{\partial F_{\mrm{post}}(\Theta_{\mrm{old}})}{\partial \sigma_{\mrm{old}}^2}
- \frac{n K}{2\sigma^2},
\label{eqn:grad-sigma_freeEnergyExpression}
\end{align}
and
\begin{align}
\nabla_{\alpha}Q(\Theta ; \Theta_{\mrm{old}}) =\frac{\partial F_{\mrm{pri}}(\theta)}{\partial \alpha} - \frac{\partial F_{\mrm{post}}(\Theta_{\mrm{old}})}{\partial \alpha_{\mrm{old}}}.
\label{eqn:grad-alpha_freeEnergyExpression}
\end{align}
In the following sections, we analyze the two free energies, $F_{\mrm{pri}}(\theta)$ and $F_{\mrm{post}}(\Theta)$, and their derivatives.

\subsection{Prior Free Energy and Its Derivatives} 
\label{sec:FreeEnergy_prior}

From Eq. (\ref{eqn:partition-function_Mat_prior}), the free energy of the prior distribution in Eq. (\ref{eqn:def_free-energy_prior}) is expressed as
\begin{align}
F_{\mrm{pri}}(\theta)
= - \frac{1}{2}  \ve{b}^{\mrm{t}} \ve{S}_{\mrm{pri}}^{-1} \ve{b} + \frac{1}{2} \ln \det \ve{S}_{\mrm{pri}} - \frac{n}{2}\ln (2\pi).
\label{eqn:free-energy_prior}
\end{align}
Since the GMRF model is defined on a $v \times v$ square grid graph $G(V, E)$ with the free boundary condition, 
its graph Laplacian $\ve{\Lambda}$ can be diagonalized as~\cite{SDCT2015,SDCT2017,SDCT2017-Okada}
\begin{align}
\ve{\Lambda} = \ve{U} \ve{\Phi} \ve{U}^{\mrm{t}}.
\label{eqn:Lambda_diagonalized}
\end{align}
Here, $\ve{U}$ is the orthogonal matrix defined by
\begin{align}
 \ve{U}:= \ve{K} \otimes \ve{K},
\label{eqn:def_matrixU}
\end{align}
where $\ve{K} \in \mathbb{R}^{v \times v}$ is a (type-II) inverse discrete cosine transform (IDCT) matrix $\ve{K} \in \mathbb{R}^{v \times v}$:
\begin{align}
K_{i,j} := 
\begin{cases}
\frac{1}{\sqrt{v}} & j = 1\\
\sqrt{\frac{2}{v}}\cos\Big\{ \frac{\pi (j-1)}{v}\Big(i - \frac{1}{2}\Big)\} & j\not=1
\end{cases}
.
\label{eqn:def_matrixK}
\end{align}
$\ve{\Phi}$ is the diagonal matrix, the elements of which are the eigenvalues of $\ve{\Lambda}$, 
and its diagonal vector, $(\Phi_1, \Phi_2,\ldots, \Phi_n)^{\mrm{t}}$, is the vector obtained by the row-major-order vectorization of the $v \times v$ matrix defined as   
\begin{align}
\psi_{i,j} := 4 \sin^2\Big(\frac{\pi (i-1)}{2 v}\Big) + 4 \sin^2\Big(\frac{\pi (j-1)}{2 v}\Big),
\label{eqn:eigenvalue_GraphLaplacian}
\end{align}
i.e., $\Phi_{i + (j - 1)v} = \psi_{i,j}$.
By using Eq. (\ref{eqn:Lambda_diagonalized}), Eq. (\ref{eqn:free-energy_prior}) is rewritten as
\begin{align}
F_{\mrm{pri}}(\theta)
= - \frac{1}{2}  \frac{n b^2}{\lambda} + \frac{1}{2} \sum_{i \in V}\ln ( \lambda + \alpha \Phi_{i} ) - \frac{n}{2}\ln (2\pi).
\label{eqn:free-energy_prior_trans}
\end{align}
The detailed derivation of this equation is shown in \ref{app:Derivation_prior_freeEnergy}.

From Eq. (\ref{eqn:free-energy_prior_trans}),
\begin{align}
\frac{\partial F_{\mrm{pri}}(\theta)}{\partial b}&=-\frac{n b}{\lambda}
\label{eqn:grad-b_free-energy_pri}\\
\frac{\partial F_{\mrm{pri}}(\theta)}{\partial \lambda} &= \frac{n b^2}{2 \lambda^2}
+\frac{1}{2}\sum_{i \in V}\frac{1}{\lambda + \alpha \Phi_{i}}, 
\label{eqn:grad-lambda_free-energy_pri}\\
\frac{\partial F_{\mrm{pri}}(\theta)}{\partial \alpha} &= \frac{1}{2}\sum_{i\in V}\frac{\Phi_{i}}{\lambda + \alpha \Phi_{i}}
\label{eqn:grad-alpha_free-energy_pri}
\end{align}
are obtained.
We can compute the right hand sides of Eqs. (\ref{eqn:grad-b_free-energy_pri})--(\ref{eqn:grad-alpha_free-energy_pri}) in $O(n)$-time.

\subsection{Posterior Free Energy and Its Derivatives} 
\label{sec:FreeEnergy_posterior}

From Eqs. (\ref{eqn:partition-function_Mat_post}) and (\ref{eqn:def_free-energy_post}), the free energy of the posterior distribution is expressed as
\begin{align}
F_{\mrm{post}}(\Theta)
&= \frac{1}{2\sigma^2}\sum_{k = 1}^K ||\ve{y}^{(k)}||^2 - \frac{1}{2}  \ve{c}^{\mrm{t}} \ve{S}_{\mrm{post}}^{-1} \ve{c}\nn
\aleq
+ \frac{1}{2}\ln \det \ve{S}_{\mrm{post}}- \frac{n}{2}\ln (2\pi).
\label{eqn:free-energy_post}
\end{align}
Because $\det \ve{S}_{\mrm{post}} = \prod_{i \in V}(\lambda + K / \sigma^2+ \alpha \Phi_{i})$, 
it can be rewritten as
\begin{align}
F_{\mrm{post}}(\Theta) &= \frac{1}{2\sigma^2}\sum_{k = 1}^K ||\ve{y}^{(k)}||^2 - \frac{1}{2}  \ve{c}^{\mrm{t}} \ve{S}_{\mrm{post}}^{-1} \ve{c}\nn
\aleq
+ \frac{1}{2}\sum_{i\in V} \ln \Big(\lambda + \frac{K}{\sigma^2} + \alpha \Phi_{i} \Big)- \frac{n}{2}\ln (2\pi).
\label{eqn:free-energy_post_trans}
\end{align}

From the derivation presented in \ref{app:Derivation_grad_b_posterior_freeEnergy}, we obtain
\begin{align}
\frac{\partial F_{\mrm{post}}(\Theta)}{\partial b} = - \frac{ n b + n(K / \sigma^2) \hat{y}_{\mrm{ave}}}{\lambda + K / \sigma^2},
\label{eqn:grad-b_free-energy_post}
\end{align}
where $\hat{y}_{\mrm{ave}} := \sum_{i \in V} \hat{y}_i / n$ is the average intensity over $\hat{\ve{y}}$.
By using the two equations   
\begin{align*}
\frac{\partial \ve{S}_{\mrm{post}}^{-1}}{\partial \gamma} = - \ve{S}_{\mrm{post}}^{-1} \frac{\partial \ve{S}_{\mrm{post}}}{\partial \gamma}\ve{S}_{\mrm{post}}^{-1},
\end{align*}
for $\gamma \in \{\lambda, \alpha, \sigma^2\}$, and $\ve{m}(Y_K, \Theta) = \ve{S}_{\mrm{post}}^{-1} \ve{c}$, we obtain the derivatives of Eq. (\ref{eqn:free-energy_post_trans}) as
\begin{align}
\frac{\partial F_{\mrm{post}}(\Theta)}{\partial \lambda} &= \frac{1}{2} || \ve{m}(Y_K, \Theta)||^2 \nn
\aleq
+\frac{1}{2}\sum_{i \in V} \frac{1}{\lambda + K / \sigma^2+ \alpha \Phi_{i}}, 
\label{eqn:grad-lambda_free-energy_post}\\
\frac{\partial F_{\mrm{post}}(\Theta)}{\partial \sigma^2} &= -\frac{1}{2\sigma^4} \sum_{k = 1}^K|| \ve{y}^{(k)} - \ve{m}(Y_K, \Theta)||^2 \nn
\aleq
-\frac{K}{2 \sigma^4}\sum_{i\in V}\frac{1}{\lambda + K / \sigma^2+ \alpha \Phi_{i}}, 
\label{eqn:grad-sigma_free-energy_post}\\
\frac{\partial F_{\mrm{post}}(\Theta)}{\partial \alpha} &= \frac{1}{2}\sum_{\{i,j\} \in E} \big(m_i(Y_K, \Theta) - m_j(Y_K, \Theta) \big)^2\nn
\aleq
+\frac{1}{2}\sum_{i \in V}\frac{\Phi_{i}}{\lambda + K / \sigma^2+ \alpha \Phi_{i}}.
\label{eqn:grad-alpha_free-energy_post}
\end{align}
Since Eq. (\ref{eqn:GS-method_post}) can be solved in $O(n)$-time, 
we can also compute the right hand side of Eqs. (\ref{eqn:grad-lambda_free-energy_post})--(\ref{eqn:grad-alpha_free-energy_post}) in $O(n)$-time. 
Note that, since $|E| < 4n$, one can compute the first term in the right hand side of Eq. (\ref{eqn:grad-alpha_free-energy_post}) in $O(n)$-time.

\subsection{Proposed Method}

From the results in Sects. \ref{sec:FreeEnergy_prior} and \ref{sec:FreeEnergy_posterior}, 
the gradients of the $Q$-function in Eqs. (\ref{eqn:grad-b_freeEnergyExpression})--(\ref{eqn:grad-alpha_freeEnergyExpression}) can be rewritten as follows.
From Eqs. (\ref{eqn:grad-b_freeEnergyExpression}), (\ref{eqn:grad-b_free-energy_pri}), and (\ref{eqn:grad-b_free-energy_post}), we obtain
\begin{align}
\nabla_b Q(\Theta ; \Theta_{\mrm{old}}) =  \frac{n b_{\mrm{old}} + n(K / \sigma_{\mrm{old}}^2) \hat{y}_{\mrm{ave}}}{\lambda_{\mrm{old}} + K / \sigma_{\mrm{old}}^2} - \frac{n b}{\lambda}.
\label{eqn:grad-b_linear}
\end{align}
From Eqs. (\ref{eqn:grad-lambda_freeEnergyExpression}), (\ref{eqn:grad-lambda_free-energy_pri}), and (\ref{eqn:grad-lambda_free-energy_post}), we obtain
\begin{align}
&\nabla_{\lambda} Q(\Theta ; \Theta_{\mrm{old}}) =- \frac{1}{2} || \ve{m}(Y_K, \Theta_{\mrm{old}})||^2 
 \nn
&-\frac{1}{2}\sum_{i \in V}\frac{1}{\lambda_{\mrm{old}} + K / \sigma_{\mrm{old}}^2+ \alpha_{\mrm{old}} \Phi_{i}}
+\frac{nb^2}{2 \lambda^2}\nn
&
+\frac{1}{2}\sum_{i \in V}\frac{1}{\lambda + \alpha \Phi_{i}}.
\label{eqn:grad-lambda_linear}
\end{align}
From Eqs. (\ref{eqn:grad-sigma_freeEnergyExpression}) and (\ref{eqn:grad-sigma_free-energy_post}), we obtain
\begin{align}
&\nabla_{\sigma^2} Q(\Theta ; \Theta_{\mrm{old}}) = \frac{1}{2 \sigma^4} \sum_{k = 1}^K || \ve{m}(Y_K, \Theta_{\mrm{old}}) - \ve{y}^{(k)} ||^2\nn
&
+ \frac{K}{2 \sigma^4} \sum_{i \in V} \frac{1}{\lambda_{\mrm{old}} + K / \sigma_{\mrm{old}}^2+ \alpha_{\mrm{old}} \Phi_{i}}
- \frac{n K}{2\sigma^2}.
\label{eqn:grad-sigma_linear}
\end{align}
Finally, from Eqs. (\ref{eqn:grad-alpha_freeEnergyExpression}), (\ref{eqn:grad-alpha_free-energy_pri}), and (\ref{eqn:grad-alpha_free-energy_post}), we obtain
\begin{align}
&\nabla_{\alpha} Q(\Theta ; \Theta_{\mrm{old}}) \nn
&=-  \frac{1}{2}\sum_{\{i,j\} \in E} (m_i(Y_K, \Theta_{\mrm{old}}) - m_j(Y_K, \Theta_{\mrm{old}}))^2 \nn
&-\frac{1}{2}\sum_{i \in V}\frac{\Phi_{i}}{\lambda_{\mrm{old}} + K / \sigma_{\mrm{old}}^2+ \alpha_{\mrm{old}} \Phi_{i}}
+\frac{1}{2}\sum_{i \in V}\frac{\Phi_{i}}{\lambda + \alpha \Phi_{i}}.
\label{eqn:grad-alpha_linear}
\end{align}
Note that $\ve{m}(Y_K, \Theta_{\mrm{old}})$ in Eqs. (\ref{eqn:grad-lambda_linear})--(\ref{eqn:grad-alpha_linear}) is the mean-vector of $P_{\mrm{post}}(\ve{x} \mid Y_K, \Theta_{\mrm{old}})$ 
and is the solution to the mean-field equation in Eq. (\ref{eqn:GS-method_post}).
Because the gradients are zero at the maximum point of $Q(\Theta ; \Theta_{\mrm{old}})$, Eq. (\ref{eqn:grad-sigma_linear}) becomes 
\begin{align}
\sigma^2 &= \frac{1}{n K}\sum_{k = 1}^K || \ve{m}(Y_K, \Theta_{\mrm{old}}) - \ve{y}^{(k)} ||^2 \nn
\aleq
+ \frac{1}{n} \sum_{i \in V} \frac{1}{\lambda_{\mrm{old}} + K / \sigma_{\mrm{old}}^2+ \alpha_{\mrm{old}} \Phi_{i}}
\label{eqn:grad-sigma_linear_trans}
\end{align}
at that point. 
Since $\ve{m}(Y_K, \Theta_{\mrm{old}})$ can be obtained in a linear time, 
we also obtain the right hand side of Eqs. (\ref{eqn:grad-lambda_linear})--(\ref{eqn:grad-sigma_linear_trans}) in a linear time.

\begin{algorithm}[H]
\caption{Linear-Time EM Algorithm for GMRF}
\label{alg:LinearTimeEM}
\begin{algorithmic}[1]
\State \textbf{Input} $Y_K$ 
\State Initialize $\Theta_{\mrm{old}}$, and $\Theta = \Theta_{\mrm{old}}$
\State Initialize $\ve{m}^{\mrm{post}}$ by the averaged image $\hat{\ve{y}}$
\Repeat
\ForAll{$t = 1,2,\ldots, T_{\mrm{mf}}$}
\ForAll{$i = 1,2,\ldots, n$}
\begin{align*}
m_i &\leftarrow  \frac{1}{\lambda_{\mrm{old}} + \alpha_{\mrm{old}} |\partial(i)| + K / \sigma_{\mrm{old}}^2}\nn
&\times
\Big( b_{\mrm{old}} + \frac{K}{\sigma_{\mrm{old}}^2}\hat{y}_i + \alpha_{\mrm{old}} \sum_{j \in \partial(i)} m_j\Big)
\end{align*}
\EndFor
\EndFor
\State Update $\sigma^2$ by using Eq. (\ref{eqn:grad-sigma_linear_trans})
\State Initialize $b$, $\lambda$, and $\alpha$
\ForAll{$t = 1,2,\ldots, T_{\mrm{M}}$}
\State $b \gets b + (\eta_b / n) \nabla_b Q(\Theta ; \Theta_{\mrm{old}})$ using Eq. (\ref{eqn:grad-b_linear})
\State $\lambda \gets \lambda + (\eta_{\lambda} / n) \nabla_{\lambda} Q(\Theta ; \Theta_{\mrm{old}})$ using Eq. (\ref{eqn:grad-lambda_linear})
\State $\alpha \gets \alpha + (\eta_{\alpha} / n) \nabla_{\lambda} Q(\Theta ; \Theta_{\mrm{old}})$ using Eq. (\ref{eqn:grad-alpha_linear})
\EndFor
\State $\Theta_{\mrm{old}} \gets \Theta$
\Until{$\Theta_{\mrm{old}} = \Theta$}
\State \textbf{Output} $\Theta$ and $\ve{m}^{\mrm{post}}$
\end{algorithmic}
\end{algorithm}
The pseudo-code of the proposed procedure is shown in Algorithm \ref{alg:LinearTimeEM}. 
The EM algorithm contains a double-loop structure: the outer loop (Steps 4 to 17) and the inner loop (Steps 11 to 15), referred to as the M-step. 
In the M-step, we maximize $Q(\Theta ; \Theta_{\mrm{old}})$ with respect to $\Theta$ for a fixed $\Theta_{\mrm{old}}$ with a gradient ascent method with $T_{\mrm{M}}$ iterations, 
where $\eta_b$, $\eta_{\lambda}$, and $\eta_{\alpha}$ are the step rates in the gradient ascent method and the gradients are presented in 
Eqs. (\ref{eqn:grad-b_linear}), (\ref{eqn:grad-lambda_linear}), and (\ref{eqn:grad-alpha_linear}).
In Steps 5 to 8, we compute the mean-vector of posterior distributions by solving the mean-field equation in Eq. (\ref{eqn:GS-method_post}) with successive iteration methods with warm restarts. 
In practice, we can stop this iteration after a few times. 
In the experiment in the following section, this iteration was stopped after just one time, i.e., $T_{\mrm{mf}} = 1$.
In Step 10, $b$, $\lambda$, and $\alpha$ are initialized by $b_{\mrm{old}}$, $\lambda_{\mrm{old}}$, and $\alpha_{\mrm{old}}$.

Since $\Phi_1 = 0$, the third term in Eq. (\ref{eqn:grad-alpha_linear}) is approximated as
\begin{align}
\frac{1}{2}\sum_{i \in V}\frac{\Phi_{i}}{\lambda + \alpha \Phi_{i}} = \frac{1}{2}\sum_{i \in V \setminus\{1\}}\frac{\Phi_{i}}{\lambda + \alpha \Phi_{i}}
\approx \frac{n - 1}{2\alpha},
\label{eqn:grad-alpha_linear_3rdTerm_approximation}
\end{align}
when $\lambda \ll 1$. 
When $\nabla_{\alpha} Q(\Theta ; \Theta_{\mrm{old}}) = 0$ and $\lambda \ll 1$, Eq. (\ref{eqn:grad-alpha_linear}) is therefore approximated as
\begin{align}
\frac{1}{\alpha} &\approx \frac{1}{n-1}\Big(\sum_{\{i,j\} \in E} (m_i(Y_K, \Theta_{\mrm{old}}) - m_j(Y_K, \Theta_{\mrm{old}}))^2 \nn
\aleq
+\sum_{i \in V}\frac{\Phi_{i}}{\lambda_{\mrm{old}} + K / \sigma_{\mrm{old}}^2+ \alpha_{\mrm{old}} \Phi_{i}}\Big).
\label{eqn:approximate_alpha}
\end{align}
Hence, if the value of $\lambda$ at the maximum point of $Q(\Theta ; \Theta_{\mrm{old}})$ is quite small, 
the value of $\alpha$ obtained at that point is close to Eq. (\ref{eqn:approximate_alpha}), and then,  
the initializing $\alpha$ by using Eq. (\ref{eqn:approximate_alpha}) in Step 10 in Algorithm \ref{alg:LinearTimeEM} could facilitate for a faster convergence, 
when the optimal $\lambda$ obtained by the EM algorithm is expected to be quite small. 
In fact, in the all of the numerical experiments in Sect. \ref{sec:NumericalExperiment}, 
the optimal values of $\lambda$ obtained by the EM algorithm were quite small ($\lambda \approx 10^{-8}$). 
We thus used this initialization method for $\alpha$ in the following numerical experiments.

When all the gradients in Eqs. (\ref{eqn:grad-b_linear}), (\ref{eqn:grad-lambda_linear}), and (\ref{eqn:grad-alpha_linear}) are zero and $\Theta_{\mrm{old}} = \Theta$, 
namely, when the EM algorithm converges, $b = \lambda \hat{y}_{\mrm{ave}}$ from Eq. (\ref{eqn:grad-b_linear}). 
This means that $b$ vanishes during the EM algorithm when $\hat{y}_{\mrm{ave}}$, which is the average intensity over the average image of $K$ degraded images, is zero.

\section{Numerical Experiments}
\label{sec:NumericalExperiment}

\begin{figure}[tb]
\centering
\includegraphics[height=3.0cm]{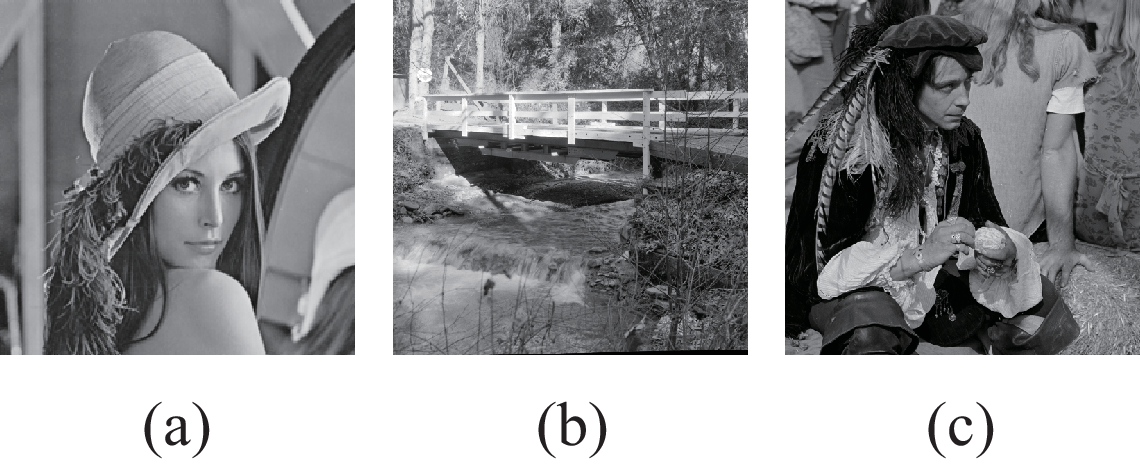}
\caption{Original 8-bit gray-scale images, the sizes of which are (a) $256 \times 256$, (b) $512 \times 512$, and (c) $1024 \times 1024$.}
\label{fig:original-images}
\end{figure}
We demonstrate our method by experiments using the images shown in Fig. \ref{fig:original-images}. 
The parameter setting of the following experiments was as follows. 
The step rates in the M-step were $\eta_b = \eta_{\alpha} = 10^{-9}$ and $\eta_{\lambda}= 10^{-13}$. 
Our Bayesian system is quite sensitive for the value of $\lambda$, and therefore we set $\eta_{\lambda}$ smaller than the others for the stability of the algorithm.
The initial values of $\Theta_{\mrm{old}}$ were $b_{\mrm{old}} = 0$, $\lambda_{\mrm{old}} = 10^{-7}$, $\alpha_{\mrm{old}} = 10^{-4}$, and $\sigma_{\mrm{old}}^2 = 2000$.
The number of iteration in the M-step, $T_{\mrm{M}}$, was one~\cite{FIT2007}, 
and the condition of convergence in Step 17 in Algorithm \ref{alg:LinearTimeEM} was
$
\max \big(| b - b_{\mrm{old}}|, | \lambda - \lambda_{\mrm{old}}|, | \alpha - \alpha_{\mrm{old}}|, | \sigma^2 - \sigma^2_{\mrm{old}}|\big) < \varepsilon,  
$
with $\varepsilon = 10^{-5}$.
The $K$ degraded images were centered (i.e., the average over all pixels in $K$ degraded images was shifted to zero) in the preprocessing. 
The maximum iteration number of the EM algorithm was 100. 

The methods, including our method, used in the following experiments were implemented with a parallelized algorithm by using C++ with OpenMP library 
for the speed-up of the for-loops (for example, the summations over $V$ and $E$ in Eqs. (\ref{eqn:grad-lambda_linear}), (\ref{eqn:grad-alpha_linear}), and (\ref{eqn:grad-sigma_linear_trans}) 
and the for-loop for $i$ in Step 6 in Algorithm \ref{alg:LinearTimeEM}),  
and they were implemented on Microsoft Windows 8 (64 bit) with Intel(R) Core(TM) i7-4930K CPU (3.4 GHz) and RAM (32 GB).

\subsection{Computation Time}
\label{sec:NumericalExperiment_Time}

We compared the computation time of our method with that of the $O(n \ln n)$-time method (the DCT--FFT method) shown in \ref{app:Nicola'sMethod}. 
The DCT--FFT method is constructed based on the DFT--FFT method proposed in Ref.~\cite{Nicolas2010}.
As shown in Tab. \ref{tab:TimeComparison}, the $O(n \ln n)$-time method was the best conventional method from the perspective of computation time.
Since the computational time of our method is $O(n)$, obviously, it is superior to the DCT--FFT method from the perspective of the order of the computational time.
However, it is important to see how fast our method is in practice as compared with the DCT--FFT method. 

In Tabs. \ref{tab:ComputationTime_s15} and \ref{tab:ComputationTime_s30}, the computational times of our method and the DCT--FFT method in specific conditions are shown. 
\begin{table}[tb]
\caption{Average computation time over 200 trials with $\sigma = 15$ and $K = 8$.}
\begin{tabular}{|c|c|c|c|} \hline
   & Fig. \ref{fig:original-images}(a) & Fig. \ref{fig:original-images}(b) & Fig. \ref{fig:original-images}(c)\\ \hline \hline
Proposed [sec] & 0.0030 & 0.0085 & 0.048  \\ \hline
DCT--FFT [sec]     & 0.0063 & 0.026 & 0.14 \\ \hline
SUR [\%]     & 51.7 \% & 67.1 \% & 65.7 \% \\ \hline
\end{tabular}
\label{tab:ComputationTime_s15}
\end{table}
\begin{table}[tb]
\caption{Average computation time over 200 trials with $\sigma = 30$ and $K = 8$.}
\begin{tabular}{|c|c|c|c|} \hline
   & Fig. \ref{fig:original-images}(a) & Fig. \ref{fig:original-images}(b) & Fig. \ref{fig:original-images}(c)\\ \hline \hline
Proposed [sec] & 0.0045 & 0.016 & 0.084  \\ \hline
DCT--FFT [sec]     & 0.0074 & 0.031 & 0.17 \\ \hline
SUR [\%]     & 40.0 \% & 48.6 \% & 49.1 \% \\ \hline
\end{tabular}
\label{tab:ComputationTime_s30}
\end{table}
The speed up rate (SUR) in the table represents the improvement rate of the computational time, which is defined as
\begin{align*} 
\mbox{SUR [\%]}:= \frac{\mbox{DCT--FFT [sec]} - \mbox{proposed [sec]}}{\mbox{DCT--FFT [sec]}} \times 100.
\end{align*}
One can observe that our method can be faster than the DCT--FFT method.

\subsection{Restoration Quality}

\begin{figure}[tb]
\centering
\includegraphics[height=3.0cm]{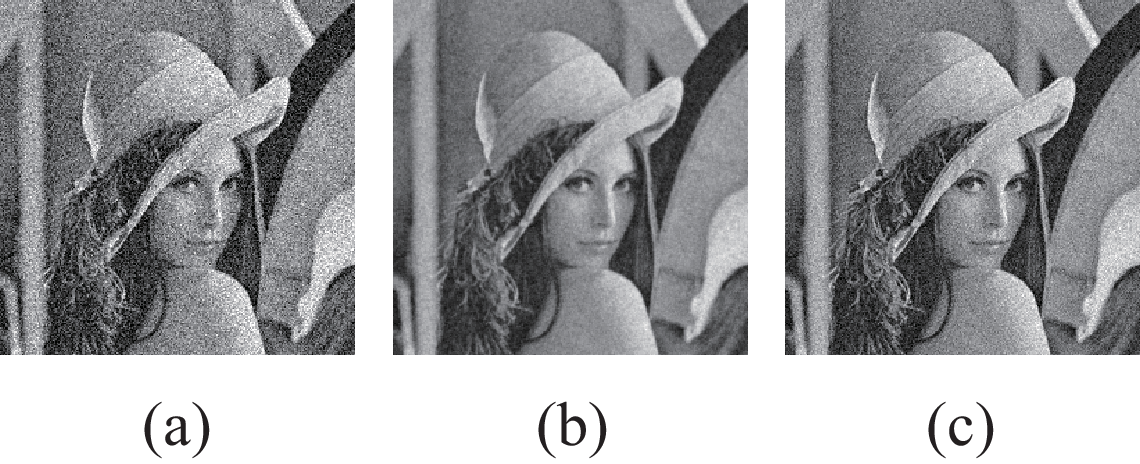}
\caption{Example of image denoising for Fig. \ref{fig:original-images}(a). (a) Example of degraded image when $\sigma = 30$ (MSE: 895.78, PSNR: 18.61), 
(b) restored image obtained by our method with $K = 3$ (MSE: 101.90, PSNR: 28.05), and (c) average image with $K = 3$ (MSE: 299.94, PSNR: 23.36).}
\label{fig:restored-images}
\end{figure}

\begin{figure*}[tb]
\centering
\includegraphics[height=4.5cm]{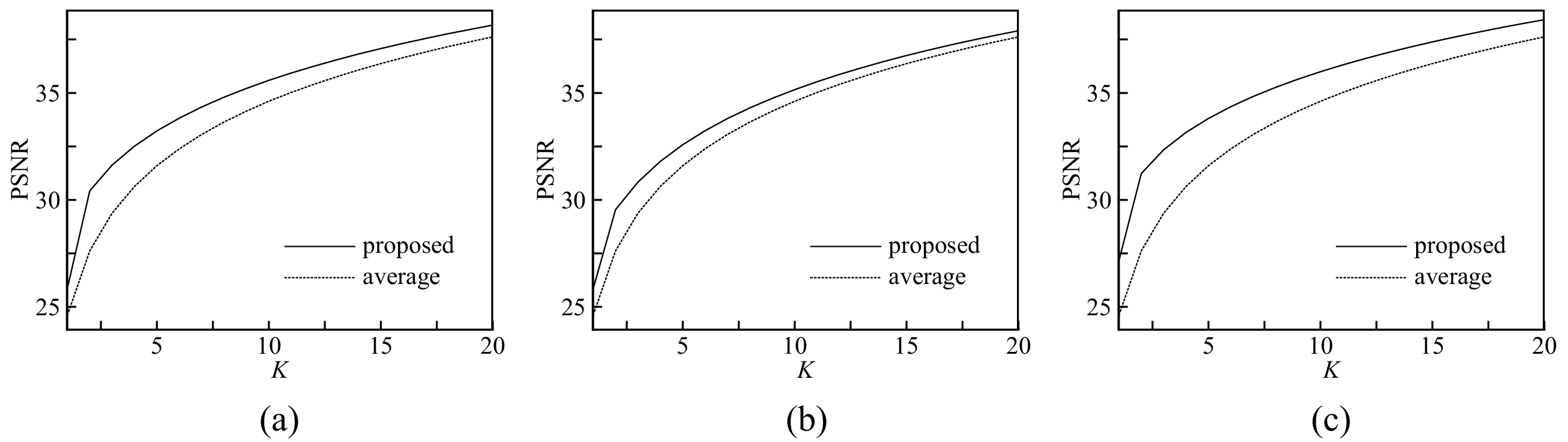}
\caption{Results of image restoration for the images shown in Figs. \ref{fig:original-images}(a)--(c) for various $K$.
The noise images were generated with $\sigma = 15$. Each plot is the average value over 200 trials.}
\label{fig:sigma15}
\end{figure*}

\begin{figure*}[tb]
\centering
\includegraphics[height=4.5cm]{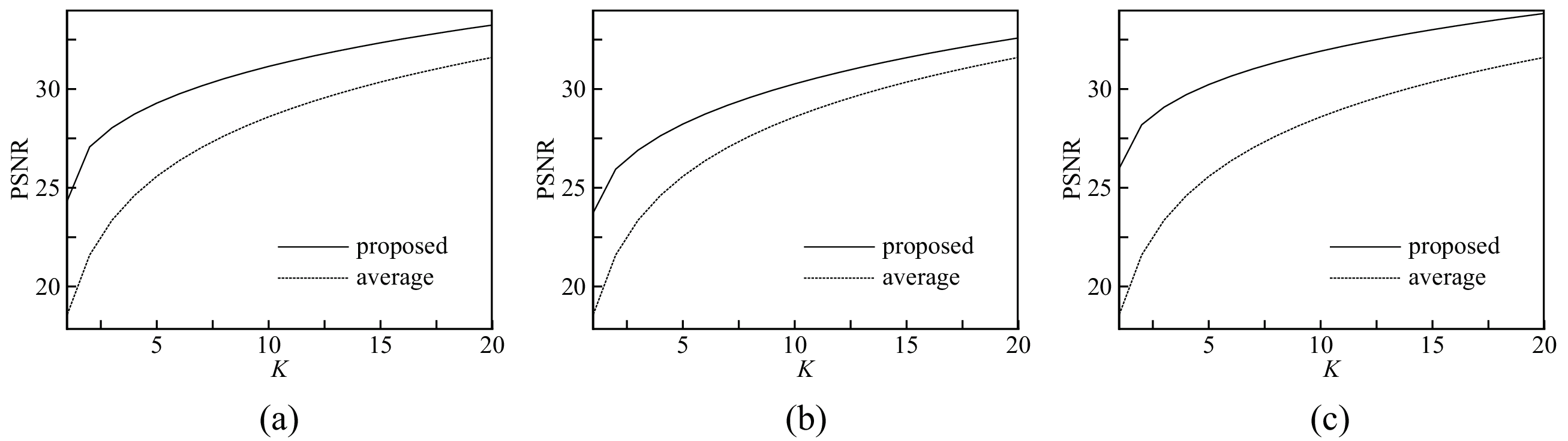}
\caption{Results of image restoration for the images shown in Figs. \ref{fig:original-images}(a)--(c) for various $K$.
The noise images were generated with $\sigma = 30$. Each plot is the average value over 200 trials.}
\label{fig:sigma30}
\end{figure*}

Here, we consider the mean square error (MSE) between the original image $\ve{x}$ and the average image of $K$ degraded images defined in Eq. (\ref{eqn:average_y_k}): 
\begin{align}
E_{\mrm{av}}:= \frac{1}{n}\sum_{i \in V}\big(\hat{y}_i - x_i \big)^2.
\label{eqn:MSE_averageImage}
\end{align}
Since $y_i^{(k)} = x_i + \nu_i^{(k)}$, where $\nu_i^{(k)}$ is the AWGN with standard deviation $\sigma$ for the $i$-th pixel in the $k$-th degraded image, 
From the law of large numbers, $E_{\mrm{av}}$ can be approximated as
\begin{align}
E_{\mrm{av}}&=\frac{1}{n}\sum_{i \in V}\Big(\frac{1}{K}\sum_{k = 1}^K \nu_i^{(k)} \Big)^2\nn
&=\frac{1}{K^2}\sum_{k,l = 1}^K\Big(\frac{1}{n} \sum_{i \in V} \nu_i^{(k)}\nu_i^{(l)}\Big) \approx \frac{\sigma^2}{K}
\label{eqn:MSE_averageImage_approx}
\end{align}
when $n \gg 1$. 
$E_{\mrm{av}}$ decreases as $K$ increases. 
Hence, one can regard the average image $\hat{\ve{y}}$ as a restored image when $K$ is large.
In the first experiment described in this section, we compared the quality of the image restoration of our method with that of the average image $\hat{\ve{y}}$.
We show an example of the denoising result when $\sigma = 30$ and $K = 3$ in Fig. \ref{fig:restored-images}.
In Figs. \ref{fig:sigma15} and \ref{fig:sigma30}, the peak signal-to-noise ratios (PSNRs) of the image restorations for $K = 1,2,\ldots, 20$ are shown. 
The noise images in Fig. \ref{fig:sigma15} were generated with $\sigma = 15$ and in Fig. \ref{fig:sigma30} with $\sigma = 30$. 
Note that the PSNR of the average image when $K = 1$ is the same as that of the noise image.
The PSNRs of the average image logarithmically grows as $K$ increases. 
This is because from Eq. (\ref{eqn:MSE_averageImage_approx}) the PSNR of the average image can be approximated as
\begin{align}
\mrm{PSNR}_{\mrm{av}} &:= 10 \log_{10} \frac{255^2}{E_{\mrm{av}}} \nn
&\approx 20 \log_{10}\frac{255}{\sigma} + 10 \log_{10}K.
\label{eqn:PSNR_averageImage_approx}
\end{align}
The results shown in Figs. \ref{fig:sigma15} and \ref{fig:sigma30} confirm that the image restored by the proposed method is always better than the average image, 
and our method is especially effective when the number of $K$ is small and the noise level is high. 

Since $E_{\mrm{av}} \to 0$ when $K \to \infty$, the average image is the perfect reconstruction of $\ve{x}$ in the limit. 
It is noteworthy that our method also has the same property, because, from Eq. (\ref{eqn:GS-method_post}), $m_i(Y_K, \Theta) = \hat{y}_i = x_i$ is obtained when $K \to \infty$ 
and all the parameters in $\Theta$ are finite.

Next, we compared the quality of image restoration of our method with that of the method with the torus approximation that is employed in the previous studies~\cite{Nishimori2000,TanakaReview2002,Nicolas2010}. 
The method with the torus approximation can be obtained by using DFT instead of DCT to diagonalize the graph Laplacian $\ve{\Lambda}$ in the derivation in \ref{app:Nicola'sMethod}.
\begin{table}[tb]
\caption{Mean square errors obtained by the proposed method and by the method with the torus approximation for Fig. \ref{fig:original-images}(a) with $\sigma = 30$.}
\begin{tabular}{|c|c|c|c|} \hline
   & $K = 1$ & $K = 3$ & $K = 5$\\ \hline \hline
Proposed & 238.9 & 102.2 & 76.7  \\ \hline
Torus approximation     & 253.4 & 105.4 & 78.8 \\ \hline
IR [\%]    & 5.7 \%& 3.1 \%& 2.7 \%\\ \hline
\end{tabular}
\label{tab:Comparision_previous(a)}
\end{table}
\begin{table}[tb]
\caption{Mean square errors obtained by the proposed method and by the method with the torus approximation for Fig. \ref{fig:original-images}(c) with $\sigma = 30$.}
\begin{tabular}{|c|c|c|c|} \hline
   & $K = 1$ & $K = 3$ & $K = 5$\\ \hline \hline
Proposed & 163.6 & 80.4 & 61.7  \\ \hline
Torus approximation     & 170.9& 82.0 & 62.8 \\ \hline
IR [\%]    & 4.2 \%& 2.0 \%& 1.7 \%\\ \hline
\end{tabular}
\label{tab:Comparision_previous(c)}
\end{table}
Tabs. \ref{tab:Comparision_previous(a)} and \ref{tab:Comparision_previous(c)} show the MSEs obtained by our method 
and by the method with the torus approximation for the images in Fig. \ref{fig:original-images}(a) and (c). 
The improving rate (IR) in Tabs. \ref{tab:Comparision_previous(a)} and \ref{tab:Comparision_previous(c)} is defined as
\begin{align*} 
&\mbox{IR [\%]}\nn
&:= \frac{\mbox{(MSE of previous method)} - \mbox{(MSE of ours)}}{\mbox{(MSE of previous method)}}\nn
\aleq \times 100.
\end{align*}
The results of our method are slightly better than those of the method with the torus approximation and the difference between the results of the two methods decreases as $K$ increases. 

In Ref.~\cite{SDCT2017-Okada}, by using synthetic data, the authors showed that denoising results obtained by a GMRF model without the PBC (i.e., our GMRF) are better 
than those obtained by a GMRF model with the PBC when data $\ve{x}$ are generated by a GMRF model without the periodic boundary condition. 
With this argument, we conclude that the prior distribution of natural images is closer to a GMRF model without the PBC than a GMRF model with it.
This conclusion is consistent with our common knowledge about natural images, namely, that there are no strong correlations among spatially distant pixels. 
Furthermore, one can observe that the difference in Tab. \ref{tab:Comparision_previous(a)} is smaller than that in Tab. \ref{tab:Comparision_previous(c)}. 
This means that the effect of the PBC decreases as the size of the image increases. 
This is consistent with the statement in Ref.~\cite{SDCT2017-Okada}.

\section{Conclusion}
\label{sec:conclusion}

In this paper, we defined a GMRF model for the Bayesian image denoising problem, 
and proposed an algorithm for solving the problem in $O(n)$-time by using DCT and the mean-field method. 
Our algorithm is state-of-the-art from the perspective of the order of the computational time. 
Moreover, our method does not need to employ torus approximation that was employed in the previous studies~\cite{Nishimori2000,TanakaReview2002,Nicolas2010}. 
By the results of the numerical experiments described in Sect. \ref{sec:NumericalExperiment}, 
we showed that our method can be faster than the $O(n \ln n)$ algorithm using FFT in practice (cf. Tabs. \ref{tab:ComputationTime_s15} and \ref{tab:ComputationTime_s30}), 
and that it can slightly outperform the method that employs torus approximation from the perspective of the restoration quality (cf. Tabs. \ref{tab:Comparision_previous(a)} and \ref{tab:Comparision_previous(c)}). 
In Figs. \ref{fig:sigma15} and \ref{fig:sigma30}, 
one can observe that the Bayesian image denoising model based on a GMRF model, considered in this paper, is more effective when the noise level is higher. 

\appendix

\section{Mean-Field Approximation for GMRF}
\label{appendix:MFE}

In this section, we briefly discuss the mean-field approximation for a GMRF model that was presented in Ref.~\cite{GraphicalModel2008}.
Let us consider a GMRF model  for $\ve{x} \in \mathbb{R}^n$ expressed as
\begin{align}
P(\ve{x}) \propto \exp\Big(\ve{r}^{\mrm{t}}\ve{x} - \frac{1}{2} \ve{x}^{\mrm{t}} \ve{A}\ve{x}\Big),
\label{eqn:appendix:GMRF}
\end{align}
where $\ve{r} \in \mathbb{R}^n$ and $\ve{A} \in \mathbb{R}^{n \times n}$ is a symmetric and positive definite matrix. 
The mean-field distribution for the GMRF model is obtained by minimizing the Kullback-Leibler divergence:  
\begin{align}
D[Q]:= \int_{-\infty}^{\infty} Q(\ve{x}) \ln \frac{Q(\ve{x})}{P(\ve{x})} d\ve{x},
\label{eqn:appendix:KLD}
\end{align}
where $Q(\ve{x}):= \prod_{i = 1}^n Q_i(x_i)$ is a test factorized distribution that satisfies the normalizing condition of $\int_{-\infty}^{\infty} Q(\ve{x}) d\ve{x} = 1$. 
The conditional minimization of Eq. (\ref{eqn:appendix:KLD}) with respect to $Q_i(x_i)$ results in
\begin{align*}
Q_i(x_i) \propto \exp\Big(-\frac{(x_i - m_i)^2}{2 A_{i,i}^{-1}}\Big),
\end{align*}
where $\ve{m} = (m_1, m_2,\ldots, m_n)^{\mrm{t}}$ are the solutions to the mean-field equation:
\begin{align}
m_i = \frac{1}{A_{i,i}}\Big(r_i - \sum_{j = 1; j \not=i}^n A_{i,j} m_j\Big),
\label{eqn:appendix:MeanFieldEquation}
\end{align}
where $\sum_{j = 1; j \not=i}^n$ is the sum from $j = 1$ to $j = n$, except $j = i$.
The solution to the mean-field equation is equivalent to the exact mean-vector of Eq. (\ref{eqn:appendix:GMRF}), 
because Eq. (\ref{eqn:appendix:MeanFieldEquation}) can be rewritten as $\ve{m} = \ve{A}^{-1}\ve{r}$. 
Eq. (\ref{eqn:appendix:MeanFieldEquation}) is also known as the Gauss-Seidel method or Jacobi method. 
The same equation can be obtained by using loopy belief propagation~\cite{GaBP2001,GaBP2008}.

\section{Derivation of Eq. (\ref{eqn:free-energy_prior_trans})}
\label{app:Derivation_prior_freeEnergy}

The orthogonal matrix $\ve{K}$, defined in Eq. (\ref{eqn:def_matrixK}), satisfies the relation 
\begin{align}
\sum_{i = 1}^v K_{i,k} K_{i,l}  = \delta(k,l), 
\label{eqn:orthogonality_of_K}
\end{align}
where $\delta(k,l)$ is the Kronecker delta function. 
Since $K_{i,1} = 1 / \sqrt{v}$ for all $i$, from Eq. (\ref{eqn:orthogonality_of_K}),
\begin{align}
\sum_{i = 1}^v K_{i,j} = \sqrt{v} \delta(j,1)
\label{eqn:sum_of_K}
\end{align}
is obtained.
From Eq. (\ref{eqn:sum_of_K}), we obtain
\begin{align}
\ve{K}^{\mrm{t}} \ve{1}_v = \sqrt{v}(1, 0, 0, \ldots, 0)^{\mrm{t}},
\label{eqn:K^t_prod_1}
\end{align}
where $\ve{1}_a \in \mathbb{R}^{a}$ is the $a$-dimensional all-one vector: $\ve{1}_a = (1, 1, \ldots, 1)^{\mrm{t}}$. 
From Eqs. (\ref{eqn:def_matrixU}) and (\ref{eqn:K^t_prod_1}), we obtain
\begin{align}
\ve{U}^{\mrm{t}} \ve{1}_n = \sqrt{n}(1, 0, 0, \ldots, 0)^{\mrm{t}}.
\label{eqn:U^t_prod_1}
\end{align}
By using Eqs. (\ref{eqn:precision-Mat_prior}) and (\ref{eqn:Lambda_diagonalized}) and the relation $\ve{U}^{-1} = \ve{U}^{\mrm{t}}$, 
\begin{align}
\ve{b}^{\mrm{t}} \ve{S}_{\mrm{pri}}^{-1} \ve{b} &= b^2 \ve{1}_n^{\mrm{t}}\big( \lambda \ve{I}_n + \alpha \ve{U} \ve{\Phi}\ve{U}^{\mrm{t}}\big)^{-1} \ve{1}_n \nn
&= b^2 \big(\ve{U}^{\mrm{t}} \ve{1}_n\big)^{\mrm{t}}\big( \lambda \ve{I}_n + \alpha \ve{\Phi}\big)^{-1} \ve{U}^{\mrm{t}} \ve{1}_n 
\end{align}
is obtained. By substituting Eq. (\ref{eqn:U^t_prod_1}) into this equation, since $\Phi_1 = 0$, we obtain
\begin{align}
\ve{b}^{\mrm{t}} \ve{S}_{\mrm{pri}}^{-1} \ve{b} =\frac{n b^2}{\lambda + \alpha \Phi_{1}}
=\frac{n b^2}{\lambda}.
\label{eqn:quadraticForm_prior}
\end{align}
From this equation and the relation $\det \ve{S}_{\mrm{pri}} = \prod_{i \in V}(\lambda + \alpha \Phi_{i})$, 
we obtain Eq. (\ref{eqn:free-energy_prior_trans}).

\section{Derivation of Eq. (\ref{eqn:grad-b_free-energy_post})}
\label{app:Derivation_grad_b_posterior_freeEnergy}

By using the diagonalization in Eq. (\ref{eqn:Lambda_diagonalized}), the second term in the posterior free energy in Eq. (\ref{eqn:free-energy_post_trans}) can be rewritten as
\begin{align}
&-\frac{1}{2}  \ve{c}^{\mrm{t}} \ve{S}_{\mrm{post}}^{-1} \ve{c} =- \frac{1}{2}\Big( b\ve{U}^{\mrm{t}} \ve{1}_n+ \frac{K}{\sigma^2}\ve{z}\Big)^{\mrm{t}}\nn
&\times \Big\{ \Big( \lambda + \frac{K}{\sigma^2}\Big) \ve{I}_n + \alpha \ve{\Phi}\Big\}^{-1} \Big(b\ve{U}^{\mrm{t}} \ve{1}_n+ \frac{K}{\sigma^2}\ve{z}\Big),
\label{eqn:quadraticForm_posterior}
\end{align}
where $\ve{z} := \ve{U}^{\mrm{t}}\hat{\ve{y}} \in \mathbb{R}^n$ is the DCT of the average image of $K$ degraded images defined in Eq. (\ref{eqn:average_y_k}).
From Eqs. (\ref{eqn:U^t_prod_1}) and (\ref{eqn:quadraticForm_posterior}), the derivative of the posterior free energy with respect to $b$ is
\begin{align}
\frac{\partial F_{\mrm{post}}(\Theta)}{\partial b}&=- \frac{nb + \sqrt{n}(K / \sigma^2) z_1}{\lambda + K / \sigma^2 + \alpha \Phi_{1}} \nn
&= - \frac{nb + \sqrt{n}(K / \sigma^2) z_1}{\lambda + K / \sigma^2 }.
\end{align}
Here, since $U_{i,1} = 1 / \sqrt{n}$, 
\begin{align}
z_1 = \sum_{i = 1}^n U_{i,1} \hat{y}_i = \frac{1}{\sqrt{n}} \sum_{i = 1}^n  \hat{y}_i.
\end{align}
From the above two equations, we obtain Eq. (\ref{eqn:grad-b_free-energy_post}).

\section{DCT--FFT Method}
\label{app:Nicola'sMethod}

Here, we briefly explain a key idea of the DCT--FFT method, i.e., the $O(n \ln n)$-time method. 
The DCT--FFT method is a modification of the DFT--FFT method proposed in Ref.~\cite{Nicolas2010}. 
The original DFT--FFT method proposed in Ref.~\cite{Nicolas2010} is not applicable to the present problem as it is, 
because the problem setting in Ref.~\cite{Nicolas2010} is different from the presented one: 
the original DFT--FFT method employed the torus approximation to diagonalize the graph Laplacian $\ve{\Lambda}$ with DFT 
and the original DFT--FFT method treated $b$ and $\lambda$ as fixed constants. 
The DCT--FFT method is modification of the original DFT--FFT method that is applicable to the presented problem.
In the DCT--FFT method, DCT in Eq. (\ref{eqn:Lambda_diagonalized}) is used instead of DFT in order to avoid the use of the torus approximation  
and $b$ and $\lambda$ are treated as the optimizable parameters.

By using Eq. (\ref{eqn:quadraticForm_posterior}), the posterior free energy in Eq. (\ref{eqn:free-energy_post_trans}) can be rewritten as
\begin{align}
&F_{\mrm{post}}(\Theta) \nn
& = \frac{1}{2\sigma^2}\sum_{k = 1}^K ||\ve{y}^{(k)}||^2- \frac{1}{2} \sum_{i \in V}\frac{\big(\sqrt{n}b \delta(i,1) + (K/ \sigma^2) z_i\big)^2}{\lambda + K / \sigma^2+ \alpha \Phi_{i}}
 \nn
\aleq
+ \frac{1}{2}\sum_{i\in V} \ln \Big(\lambda + \frac{K}{\sigma^2} + \alpha \Phi_{i} \Big)- \frac{n}{2}\ln (2\pi).
\label{eqn:free-energy_post_trans_DCT}
\end{align}
Therefore, the derivatives of the free energy with respect to $\lambda$, $\sigma^2$, and $\alpha$ are
\begin{align}
\frac{\partial F_{\mrm{post}}(\Theta)}{\partial \lambda} &= \frac{1}{2} \sum_{i \in V} \Big(\frac{\sqrt{n}b \delta(i,1) + (K/ \sigma^2) z_i}{\lambda + K / \sigma^2+ \alpha \Phi_{i}}\Big)^2\nn
\aleq
+\frac{1}{2}\sum_{i \in V}\frac{1}{\lambda + K / \sigma^2+ \alpha \Phi_{i}}, 
\label{eqn:grad-lambda_free-energy_post_DCT}\\
\frac{\partial F_{\mrm{post}}(\Theta)}{\partial \alpha} &= \frac{1}{2} \sum_{i \in V} \Big(\frac{\sqrt{n}b \delta(i,1) + (K/ \sigma^2) z_i}{\lambda + K / \sigma^2+ \alpha \Phi_{i}}\Big)^2\Phi_i\nn
\aleq
+\frac{1}{2}\sum_{i\in V}\frac{\Phi_{i}}{\lambda + K / \sigma^2+ \alpha \Phi_{i}},
\label{eqn:grad-alpha_free-energy_post_DCT}
\end{align}
and
\begin{align}
&\frac{\partial F_{\mrm{post}}(\Theta)}{\partial \sigma^2}=
-\frac{1}{2\sigma^4}\sum_{k = 1}^K ||\ve{y}^{(k)}||^2\nn
&+\frac{K}{\sigma^4}\sum_{i \in V}\frac{\big(\sqrt{n}b \delta(i,1) + (K/ \sigma^2) z_i\big)z_i}{\lambda + K / \sigma^2+ \alpha \Phi_{i}} \nn
&-\frac{K}{2\sigma^4} \sum_{i \in V} \Big(\frac{\sqrt{n}b \delta(i,1) + (K/ \sigma^2) z_i}{\lambda + K / \sigma^2+ \alpha \Phi_{i}}\Big)^2 \nn
&-\frac{K}{2\sigma^4}\sum_{i \in V}\frac{1}{\lambda + K / \sigma^2+ \alpha \Phi_{i}}, 
\label{eqn:grad-sigma_free-energy_post_DCT}
\end{align}
respectively. 
When DCT of the average image of $K$ degraded images, $\ve{z} = \ve{U}^{\mrm{t}}\hat{\ve{y}}$, has been obtained, 
we can compute the above derivatives in $O(n)$ time. 
By using Eqs (\ref{eqn:grad-lambda_free-energy_post_DCT})--(\ref{eqn:grad-sigma_free-energy_post_DCT}) instead of Eqs. (\ref{eqn:grad-lambda_free-energy_post})--(\ref{eqn:grad-alpha_free-energy_post}), 
the EM algorithm, similar to Algorithm \ref{alg:LinearTimeEM}, can be constructed. 
It is noteworthy that the resulting EM algorithm does not need to solve the mean-field equation (in Steps 5 to 7 in Algorithm \ref{alg:LinearTimeEM}), 
because it does not need the mean vector of the posterior distribution. 
In this EM algorithm, we need to use $\ve{z}$. 
By using FFT, it can be obtained in $O(n \ln n)$ time. 
The order of the computational cost of this EM algorithm is, then, $O(n \ln n)$. 

After the EM algorithm, in order to obtain the restored image, one has to compute the mean vector of the posterior distribution (i.e., MAP estimation in Eq. (\ref{eqn:MAP_estimation})). 
The mean vector of the posterior distribution is 
\begin{align}
&\ve{m}^{\mrm{post}} = \ve{S}_{\mrm{post}}^{-1} \ve{c}\nn
&=\ve{U} \Big\{ \Big( \lambda + \frac{K}{\sigma^2}\Big) \ve{I}_n + \alpha \ve{\Phi}\Big\}^{-1}\Big(b\ve{U}^{\mrm{t}} \ve{1}_n+ \frac{K}{\sigma^2}\ve{z}\Big),
\label{eqn:meanVector_posterior_DCT}
\end{align}
and it is the IDCT of vector $\{ ( \lambda + K/ \sigma^2) \ve{I}_n + \alpha \ve{\Phi}\}^{-1}(b\ve{U}^{\mrm{t}} \ve{1}_n+ (K / \sigma^2)\ve{z})$, 
which can be obtained in $O(n \ln n)$ time by using FFT. 
In the numerical experiment in Sect. \ref{sec:NumericalExperiment_Time}, we used the FFTW library~\cite{FFTW2005} in FFT. 

\section*{acknowledgments}
This work was partially supported by JST CREST Grant Number JPMJCR1402
and by JSPS KAKENHI Grant Numbers 15K00330, 15H03699, and 15K20870.

\bibliography{citations}
\end{document}